# 基于监督式学习的全景相机与激光雷达的联合标定[*]

曹明玮[1]，杨明[2,3]，王春香[1]，钱烨强[2,3]，王冰[2,3]

(1. 上海交通大学机器人所 上海 200240； 2. 上海交通大学自动化系，上海 200240；
3. 系统控制与信息处理教育部重点实验室 上海 200240)

**摘 要**: 针对当前全景相机-激光雷达系统，传统标定方法不适合用在全景相机这种极度非线性的成像模型上，基于统计最优化方法对激光雷达线数要求较高，精密测量标定设备昂贵等问题，提出了一种基于监督式学习的全景相机-激光雷达的联合标定方法。通过设计了一种圆形标定物获取点云和全景图对应特征点，生成训练数据集；进一步地，对全景相机的成像模型进行预处理，设计了一种包含一个隐藏层的监督式BP学习网络，将传统的标定问题转化为一个多元非线性的回归优化问题；利用反向传播算法，回归出满足精度要求的旋转平移矩阵。实验结果表明，该方法能够迅速回归出标定参数，精度优于传统方法以及基于统计最优化的方法。.该方法标定精度较高，且具有自动化程度较高等特点。
**关键词**: 全景相机;激光雷达;联合标定;监督式学习;BP算法
**中图分类号**: TP751.1　　　　**文献标志码**: A　　　　**文章编号**：(作者可不填)

# Joint calibration of panoramic camera and lidar based on supervised learning

Cao Mingwei[1], Yang Ming[2,3], Wang Chunxiang[1], Qian Yeqiang[2,3], Wang Bing[2,3]

(1.Research Institute of Robotics, Shanghai Jiao Tong University, Shanghai 200240;
2. Department of Automation, Shanghai Jiao Tong University, 200240;
3. Key Laboratory of System Control and Information Processing, Ministry of Education of China, Shanghai, 200240)

**Abstract:** In view of contemporary panoramic camera-laser scanner system, the traditional calibration method is not suitable for panoramic cameras whose imaging model is extremely nonlinear. The method based on statistical optimization has the disadvantage that the requirement of the number of laser scanner's channels is relatively high. Calibration equipments with extreme accuracy for panoramic camera-laser scanner system are costly. Facing all these in the calibration of panoramic camera-laser scanner system, a method based on supervised learning is proposed. Firstly, corresponding feature points of panoramic images and point clouds are gained to generate the training dataset by designing a round calibration object. Furthermore, the traditional calibration problem is transformed into a multiple nonlinear regression optimization problem by designing a supervised learning network with preprocessing of the panoramic imaging model. Back propagation algorithm is utilized to regress the rotation and translation matrix with high accuracy. Experimental results show that this method can quickly regress the calibration parameters and the accuracy is better than the traditional calibration method and the method based on statistical optimization. The calibration accuracy of this method is really high, and it is more highly-automated.

## 0 引言

近年来，激光雷达-相机系统在许多应用中迅速涌现。激光雷达-相机系统频繁应用于各种高级辅助驾驶系统，场景建模等应用场景。这些应用都得益于将激光测距技术与现有的视觉系统相整合，从而可以更全面地了解环境的3D结构。然而，这些技术的实现都依赖于激光雷达和图像传感器之间良好的几何校准。这些几何校准的参数确定了激光雷达坐标系如何变换到图像传感器坐标系，使得整个激光雷达-相机系统可以精确的还原出环境的图像和位置信息。

激光雷达-相机系统的联合标定的研究主要分为两种：早期的基于平面靶特征方法和基于统计最优化方法。早期传统平面靶特征方法最为典型的是由文献[1]Zhang(2004)提出的基于对于多个平面棋盘格的多次观察进行联合标定的方法。后续很多文献[2-3]等基于Zhang的方法做了一些改进，但这些方法都是针对单目相机-激光雷达系统。对于线数比较密的测绘单线以及64线激光雷达，统计和信息理论能够充分利用数据的统计一致性，不需要标定物，开始广泛应用于这类相机-激光雷达系统的标定。文献[4] Mastin A,et al(2009)提出了一种基于最大化激光雷达和遥感图像互信息来进行联合标定的方法。文献[5] Gaurav,et al(2012)和文献[6] Pandey,et al (2012)将互信息方法应用于3D激光雷达和全景相机的联合标定。这些基于互信息的方法对于激光雷达的线数要求很高，而对于线数较少的激光，由于不能满足其方法的内蕴统计性对于信息量的需求，效果较差，而且对于标定场景中含有阴影，闭塞的情况会出现标定出错的情况。

综上所述，针对传统方法主要针对单目相机，对于全景相机的极度非线性成像模型适应性不高进而导致标定精度不高，基于互信息的统计信息理论方法对于激光雷达线数要求较高等问题，本文提出了一种基于监督式BP学习网络的全景相机和激光雷达的联合标定方法，该方法标定过程相对简单，对于线数较少的激光雷达也能有很好的效果，精度以及自动化程度也相对较高。

# 1 联合标定问题模型

激光雷达-全景相机系统的联合标定包含两个变换映射的过程，一是激光雷达坐标系下的点映射到全景相机坐标系下的点的线性变换过程，二是根据相机的成像模型，由相机坐标系下的三维点到二维图像平面像素的成像过程，如图1所示：

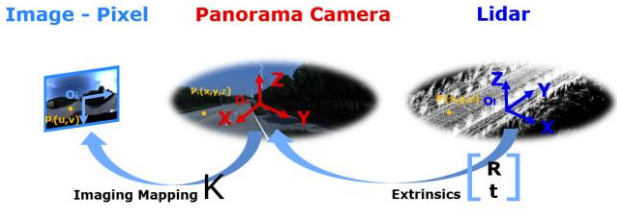

图1 激光雷达-全景相机系统的联合标定

## 1.1 激光雷达坐标系到全景相机坐标系的变换

激光雷达采集的点云是激光雷达坐标系下的点，需要经过旋转平移变换将点云转换为全景相机坐标系下的点，如图2（a），两个坐标系的变换可以表达为如下数学形式：

$$\vec{X_c} = R\vec{X_l} + T \quad (1)$$

为了后续训练的方便，将旋转矩阵表达为欧拉角的形式。欧拉角有很多不同的组合形式，而学界并没有一个共识的表述方式，因而在使用欧拉角的时候，必须要明确指出其夹角的顺序和参考轴。本文使用的zxz顺规形式的欧拉角如图2（b）所示。

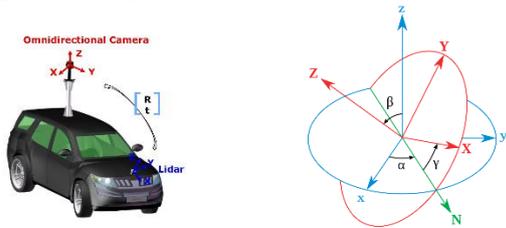

（a）坐标系变换示意图　　（b）zxz顺规的欧拉角形式
图2 坐标系变换以及zxz顺规欧拉角形式

## 1.2 球形全景相机的成像模型

球形全景相机,如图3（a），其包括周边5个及顶部1个共6个镜头。全景相机可以视作一个虚拟相机，其虚拟相机坐标系如图3（b）所示。球形全景相机有其固有成像模型，其原理示意图如图4所示。

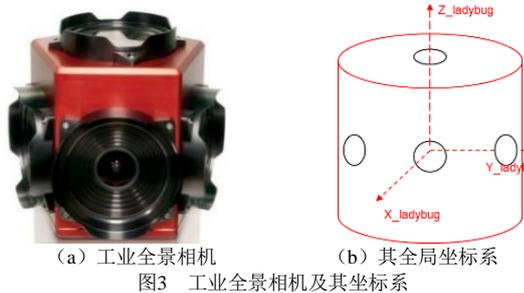

（a）工业全景相机　　（b）其全局坐标系
图3 工业全景相机及其坐标系

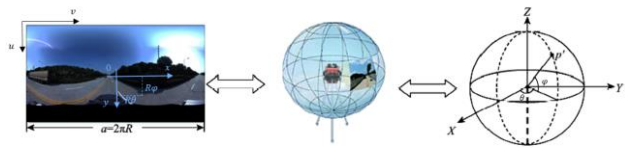

图4 球形全景图成像示意图

设球体模型上的点为$P(x_c, y_c, z_c)$，对应全景影像上的像素比例坐标$(u, v)$，其点的坐标与像素比例坐标的关系可以表示为：

$$u = \frac{\pi - \text{atan2}(y_c, x_c)}{2\pi} \quad (2)$$

$$v = \frac{\pi - 2\arctan(z_c/\sqrt{x_c^2 + y_c^2})}{2\pi} \quad (3)$$

# 2 基于监督式学习的联合标定方法

## 2.1 监督式BP学习网络设计

本文设计了一种包含一个隐藏层的监督式BP学习网络，如图5所示，网络的输入是点云的坐标即$x_l$，$y_l$，$z_l$，预期输出是等价于全景图上对应像素比例坐标$u$，$v$的$h_1$，$h_2$。

值得注意的是，本文设计的并不是一种神经网络结构，虽然现行的网络中神经网络的框架较多，搭建起来容易，但却不适用于作为本文的网络结构。

在全景成像过程即隐藏层到输出层的过程中，如图5，隐藏层的节点之间需要进行一系列非线性交互的操作，如公式(2)(3)所示。而这种全景成像内禀的非线性相互作用关系没有办法通过统一的数学处理形式消解掉，而神经网络的神经元定义如公式(4)所示，为节点之间线性加权进入激活函数的数学形式，因而在保留全景成像模型的情况下，无法应用神经网络这样一种网络结构。

$$t = f\left(\vec{W'A} + b\right) \quad (4)$$

而且就精度上考虑，本文在已知全景相机成像模型的情况下，给网络指定既定的精确的全景成像模型比放开成像这一层的约束，让神经网络自行拟合成像函数去回归标定参数精度要高。

本文监督式BP学习网络的反向传播模式如图6所示。通过不断计算每一层的梯度，将loss函数对于权值的梯度反向传播回权值，结合合适的学习率指导权值按照输出的梯度方向进行调整，从而快速迭代出loss函数极小的输入权值。

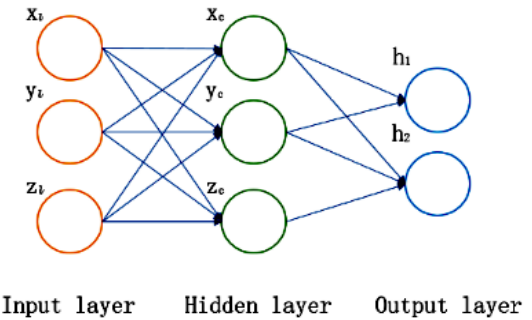

图5 监督式学习网络结构图

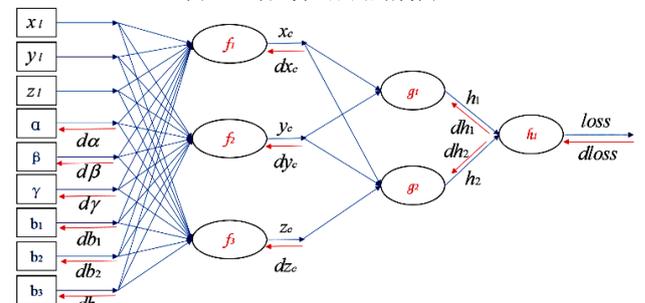

图6 监督式学习网络的计算图表示

## 2.2 模型预处理与评价函数设计

### 2.2.1 模型预处理

由于原来的全景成像模型输出过程包含atan2分段函数，如公式(2)(3)，进行反向传播推导会带来梯度问题以及非常繁杂的偏导，因而需要对模型进行一些预处理。成像模型上，取atan2函数表达式一致的部分进行标定，建立标定模型。对公式(2)(3)进行变换，将原来的模型输出($u$, $v$)进行等效调整为($h_1$, $h_2$)，变成如下(5)(6)形式，

这样可以将计算转移到评价函数里，而不用在回归过程中反复进行 arctan 复合函数复杂的偏导运算。

$$h_1 = \frac{y_c}{x_c} = \tan(\pi - 2\pi u) \quad (5)$$

$$h_2 = \frac{z_c^2}{x_c^2 + y_c^2} = (\tan(\frac{\pi}{2} - \pi v))^2 \quad (6)$$

经过模型预处理，训练模型可以表达为如下具有 6 个学习权值 $R(\alpha,\beta,\gamma)$，$T(b_1,b_2,b_3)$ 的形式：

$$\vec{X_c} = R\vec{X_l} + T \quad (7)$$

$$h_1 = \frac{y_c}{x_c} \quad (8)$$

$$h_2 = \frac{z_c^2}{x_c^2 + y_c^2} \quad (9)$$

#### 2.2.2 评价函数设计

通过对模型进行预处理，将计算转化到 loss 评价函数里面，简化了网络的计算复杂度。而在评价函数里将 ($h_1$，$h_2$) 模型输出还原为具有物理意义的像素比例 ($u$，$v$)，再与训练数据集中坐标比例的真值 ($u'$，$v'$) 作比较处理：

$$loss = \frac{1}{2}((u-u')^2 + (v-v')^2) \quad (10)$$

$$u - u' = \frac{\pi - \arctan(h_1)}{2\pi} - u'_i \quad (11)$$

$$v - v' = \frac{\frac{\pi}{2} - \arctan(\sqrt{h_2})}{\pi} - v'_i \quad (12)$$

### 2.3 训练数据集的获取

#### 2.3.1 训练数据集获取标定物设计

标定模型的训练数据集包含点云三维坐标 $x_l$，$y_l$，$z_l$ 以及对应的像素比例坐标 ($u'$，$v'$) 五列。本文设计了一种圆形标定物来获取训练数据集，如图 7 所示，显然，无论激光雷达以任何一个角度扫到实验标定物平面时，都会留下相应的点云阵列，而这些点云阵列由于长度不同，与地面的斜率以及在圆形标志牌的象限位置不同，在全景图上都有唯一的线段与之对应，从而可以根据圆几何推导点云线段端点的像素坐标。

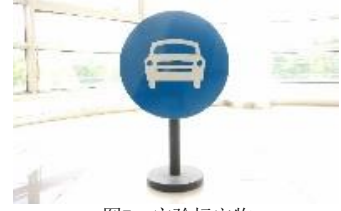

图7　实验标定物

#### 2.3.2 点云线段的分析处理

根据全景图时间戳对应相应时间帧的点云，对于每一帧点云首先使用 RANSAC 抽离出地面，其次基于 web 开源库 potree 如图 8，导出点云线段的端点坐标，线段长度，以及以垂直于地面的方向，点云线段相对于标志牌圆的几何位置，如图 9 GraphInfo 定义所示。

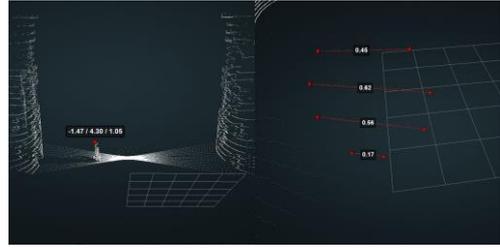

图 8　点云特征点的可视化分析工具

#### 2.3.3 全景图图像处理

同时对于每一帧的全景图像需要提取出图中标志牌的圆心坐标，方便后续根据点云线段推导的几何关系去推导点云线段的端点在全景图上对应的线段端点的像素坐标。首先对整幅图进行颜色阈值分割可以基本滤掉大部分的无关背景，然后进行高斯滤波与二值化进一步使得整幅图像只剩下标志牌的部分，这样进行霍夫圆的检测基本都能非常精确的提取出标志牌圆心的坐标。

#### 2.3.4 根据圆几何推导点云线段端点对应像素坐标

在已知了圆心像素坐标和线段相对于圆心的关系，则可通过圆几何去推导出线段端点的像素坐标，如图 9 圆几何的公式所示，从而生成点云中点的对应的像素坐标，生成后续训练所需的数据集。

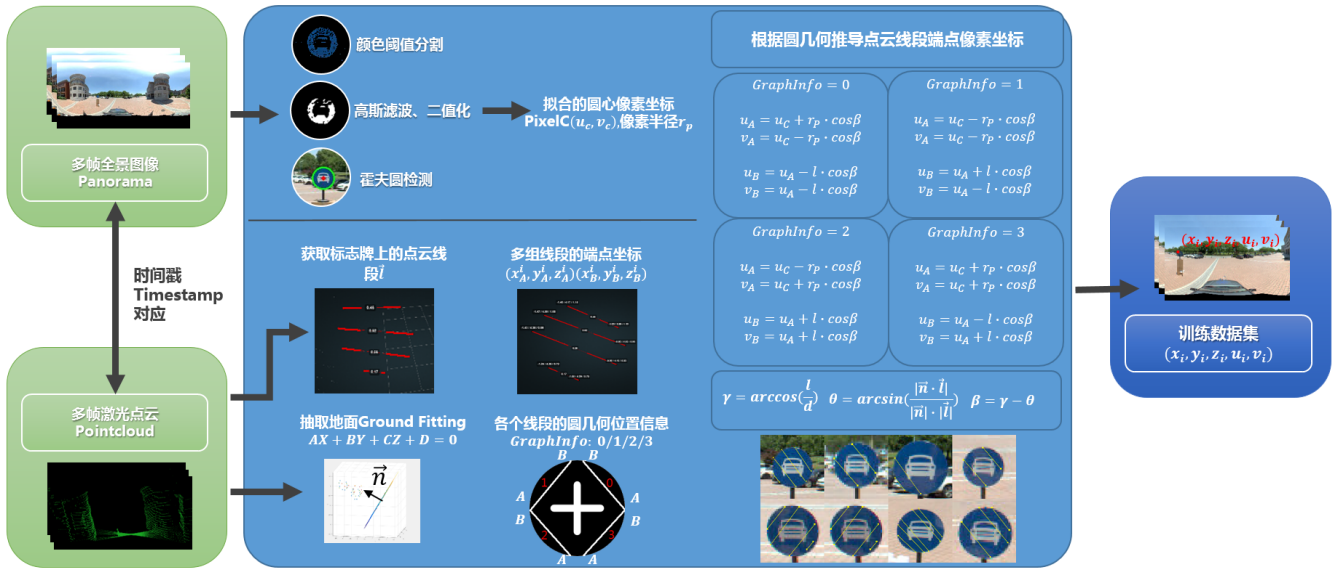

图9　根据圆几何推导点云线段端点像素坐标

# 3 实验结果与分析

## 3.1 实验平台场景与实验方法

本文以上海交通大学智能车实验室的 CyberTiggo 智能车作为实验平台，如图 10，车顶安装有工业全景相机，车前安装有 16 线激光雷达，此外实验平台还配备有双频 GPS 接收机，惯导 IMU 等感知定位传感器。本文实验场景选择在特征明显的电信楼群之间，如图 11 所示。

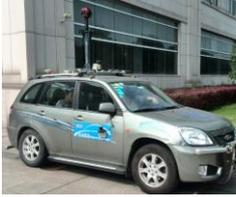 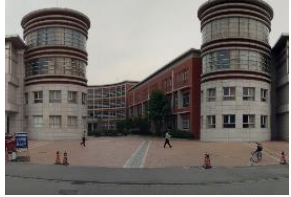

图10　CyberTiggo实验平台　　图11　联合标定实验场景

- 在车辆前进行好标定物的设置，如图12所示，让车辆带速前行，采集精确GPS数据，惯导数据，激光雷达点云数据，全景相机图像数据。

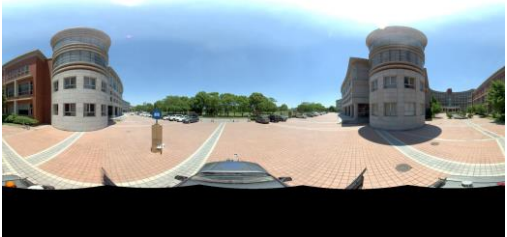

图12　实验标定物的设置

## 3.2 训练与标定结果分析

### 3.2.1 训练结果分析

由于外参旋转平移矩阵只有六个未知量，理论上最少 3 个对应特征点就可以进行求解，本文采用了 8 帧全景图与对应时间戳的 8 帧点云生成了 48 个对应点的训练数据集，实际实验数据训练中，整个训练梯度下降的过程非常的迅速，整个 20000 次看起来近乎是阶跃的如图 13(a)，即使放到前 100 次下降速度仍然是极快的，如图 13(b)。本文方法达到的旋转平移矩阵最终的精度满足实际使用需求。实际训练过程发现，较之平移矩阵，旋转矩阵的参数是更难学习的权值，如图 13（c）（d）所示。

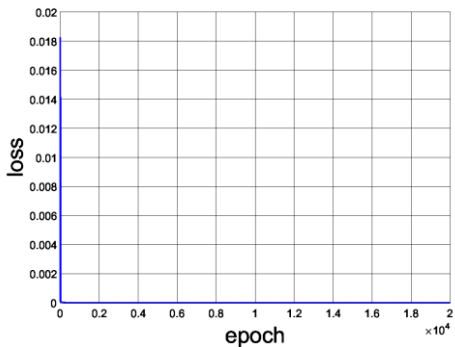

(a) loss 整个 20000 次训练

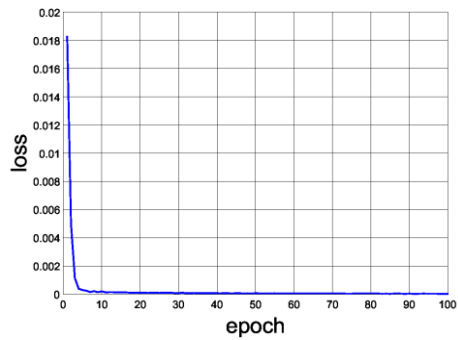

(b) loss 前 100 次训练

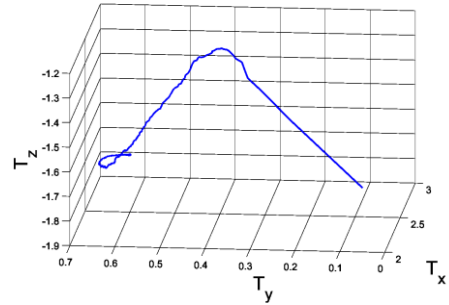

(c) 平移矩阵解的收敛性

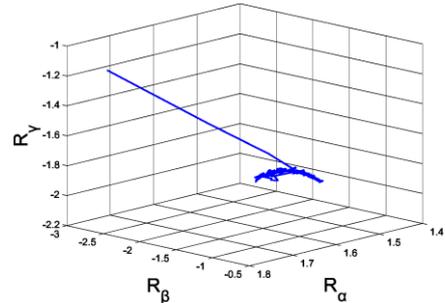

(d) 旋转矩阵解的收敛性

图13　训练结果与解的收敛性

### 3.2.2 标定结果与分析

A.点云-全景图投影结果

利用回归出的旋转平移矩阵，结合全景图的成像模型，将点云投影到全景图上，可以验证标定的精度，如图 14，可以看到点云上的点都能比较好的贴合场景中各个特征。

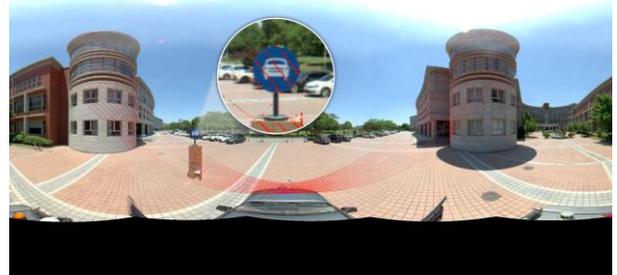

图14　点云-全景图投影结果

B.全景图-点云上色结果

精确的联合标定结果可以用于彩色点云的生成，通过将高精度的 GPS/IMU 进行数据融合，得到高精度定位，再利用高精度定位结果将采集的 16 线激光雷达的点云数据叠加起来，形成较为稠密的三维点云，利用联合标定结果将点云映射到全景图像素，并从像素上获取 RGB 值给点云上色，如图 15，进一步验证了联合标定的精度。

C.实验方法对比

本文将提出了基于监督式学习的方法与传统方法[1]以及基于互信息的统计最优化方法进行了对比，如表 1，旋转和平移矩阵在精度上均略优于传统标定方法，重投影误差更小。传统方法由于非线性的最小二乘法没有闭式解，因而一般还是用迭代法去求解，对于全景相机的适应性不高，相对还是更适合单目相机-激光雷达系统；其次，基于统计最优化的方法对于激光雷达的线数要求还是比较严格，在不满足其线数需求时，统计特性不能很好的提取出信息，因而精度不是很高。因而本文方法主要是针对成像模型高度非线性的全景相机与线数不那么稠密的激光雷达联合标定，相对更加简单，精度，自动化程度更高。

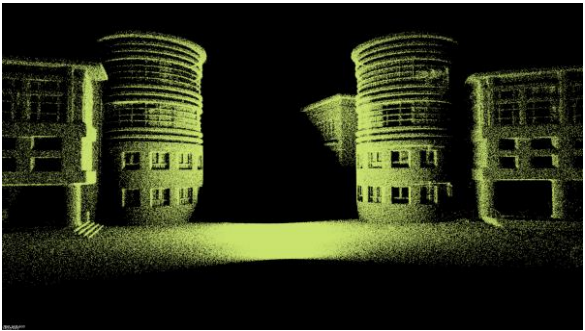
(a) 叠加后的稠密点云整体（未上色）

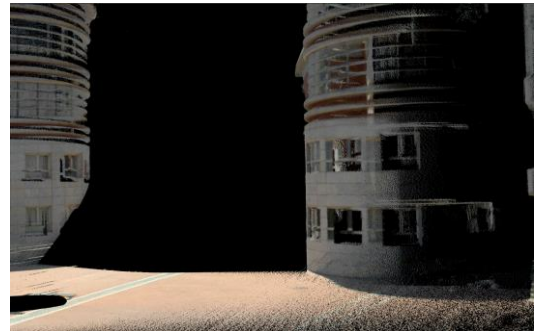
(c) 上色后的点云局部

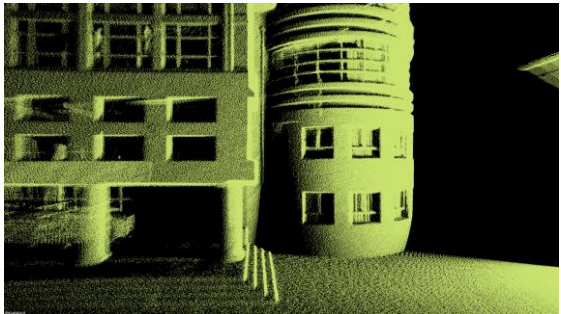
(b) 叠加后的稠密点云局部（未上色）

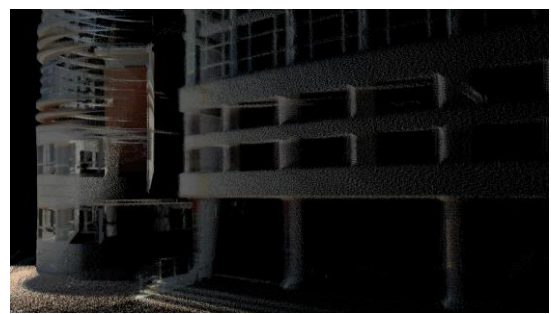
(d) 上色后的点云局部

图15 全景图-点云上色结果图

表1 本文方法与传统方法的精度对比

| Items | Rotation Matrix R[α,β, γ]/rad | Translation Vector T/m | Average Re-projection Error |
|---|---|---|---|
| Traditional Calibration Method | [4.72, 0.90, 1.82] | [2.83, 0.65, -1.76] | Vertical: 1.2 pixel Horizontal: 0.8 pixel |
| Mutual Information Method | [4.651, 0.912, 1.822] | [2.771, 0.643, -1.772] | Vertical: 2.3 pixel Horizontal: 1.6 pixel |
| Learning-method Calibration | [4.7112, 0.8932, 1.8420] | [2.8673, 0.6389, -1.7732] | Vertical: 0.7 pixel Horizontal: 0.4 pixel |

## 4 结束语

本文提出了一种针对全景相机-激光雷达系统的监督式学习标定方法，通过设计了一种包含一个隐藏层的监督式 BP 学习网络将传统的标定问题转化为一个多元非线性的回归优化问题。首先，本文设计了一种圆形的标定物获取全景图和点云的对应特征点，生成网络训练需要的数据集。其次，针对全景相机成像模型的复杂非线性的特点，对其作了模型预处理，使得网络计算的复杂度大大降低。实验结果表明，该方法标定精度较高，对于线数不那么稠密的激光雷达也能有很好的效果，且具有自动化程度较高，耗时相对较短等特点。